\documentclass[letterpaper, 10 pt, conference]{ieeeconf}  

\usepackage{multicol}
\usepackage[bookmarks=true]{hyperref}

\usepackage{amssymb}
\usepackage{amsmath}
\usepackage{float}
\usepackage{cite}
\usepackage{dblfloatfix}
\usepackage{graphicx}
\usepackage{mathtools}
\usepackage{siunitx}
\usepackage{filecontents}
\usepackage[font=footnotesize,labelfont=footnotesize]{subcaption}
\usepackage[font=footnotesize,labelfont=footnotesize]{caption}

\usepackage{comment}
\usepackage{authblk}
\usepackage{multirow}
\usepackage{environ}

\usepackage{amsthm}

\usepackage{algorithm, algorithmicx}
\usepackage[noend]{algpseudocode}
\usepackage[english]{babel}

\algnewcommand\algorithmicinput{\textbf{Input:}}
\algnewcommand\algorithmicoutput{\textbf{Output:}}
\algnewcommand\algorithmicinitialize{\textbf{Initialize:}}
\algnewcommand\Input{\item[\algorithmicinput]}%
\algnewcommand\Output{\item[\algorithmicoutput]}%
\algnewcommand\Initialize{\item[\algorithmicinitialize]}%

\DeclareMathOperator*{\argmin}{arg\,min}

\theoremstyle{plain}

\graphicspath{ {img/} }

\IEEEoverridecommandlockouts    

\title{\LARGE \bf
Configuration Control for Physical Coupling of Heterogeneous Robot Swarms
}

\author{Sha Yi \and Zeynep Temel \and Katia Sycara
\thanks{The authors are with the Robotics Institute, Carnegie Mellon University, Pittsburgh, PA 15213, USA. Email: {\tt\small \{shayi, ztemel, katia\}@cs.cmu.edu}.}
\thanks{This work was funded by AFOSR award FA9550-18-1-0097 and AFRL/AFOSR award FA9550-18-1-0251.}
\thanks{Hardware and Software implementation is available at: {\tt\url{https://github.com/ZoomLabCMU/puzzlebot_v2}}}}
\begin{document}
\maketitle
\begin{abstract}
In this paper, we present a heterogeneous robot swarm system that can physically couple with each other to form functional structures and dynamically decouple to perform individual tasks. The connection between robots can be formed with a passive coupling mechanism, ensuring minimum energy consumption during coupling and decoupling behavior. The heterogeneity of the system enables the robots to perform structural enhancement configurations based on specific environmental requirements. We propose a \textit{connection-pair oriented configuration control} algorithm to form different assemblies. We show experiments of up to nine robots performing the coupling, gap-crossing, and decoupling behaviors.
\end{abstract}

\section{Introduction}\label{sec:intro}
In unstructured environments, ants form and adapt functional structures dynamically in response to obstacles, gaps, and holes \cite{reid2015army}. Inspired by these animals, robot swarms perform collective behaviors and accomplish complex tasks that a single robot is not capable of. In this paper, we build on our previous work that introduced PuzzleBots \cite{yi2021puzzlebots} - a reconfigurable robot swarm system with a passive coupling mechanism, and present extensions on structural enhancement configuration control. 

Existing robot swarms, or Multi-Robot System (MRS) platforms \cite{pickem2017robotarium, rubenstein2012kilobot} have demonstrated collective and decentralized collaboration, but the robots do not physically interact with each other - physical abilities of the robots remain the same as a single robot. In the modular robot systems, individual units are equipped with active mechanical structures to couple and form various structures \cite{tosun2016Design, romanishin2013Mblocks, liang2020freebot, gross2006autonomous, haghighat2015lily}. The most common method for dynamic coupling is performed by magnetic forces \cite{tosun2016Design, romanishin2013Mblocks, liang2020freebot, saldana2019design}. This may consume high energy during the coupling or decoupling process, and also has limited load-carrying capabilities. In addition, the magnets are directional, limiting the formation of the robots and introducing complexity in controlling and planning algorithms. In most modular robot systems, each connection is connected via a single contact point/face. Single connection is more fragile compared with multiple connections when encountering complex environments \cite{luo2019minimum}. Reconfiguration algorithms for pre-connected modular robot systems focus on graph topology reconfiguration \cite{liu2020motion, liu2019distributed, hou2014graph}. Modular robots that have limited mobility reconfigure based on motion primitives \cite{vassilvitskii2002complete} or grid-based setup \cite{claici2017distributed, saldana2017decentralized}. These methods restrict the formation and are ineffective with systems where robots have individual mobility and no strong connection with each other.

In PuzzleBots \cite{yi2021puzzlebots}, we presented a passive coupling mechanism where each robot has knobs and holes. PuzzleBots can couple with each other by pushing the knobs into the holes of the other robot when initially aligned, cross a gap, and decouple to perform individual tasks. The passive coupling mechanism does not consume any additional power compared to active coupling mechanisms. While the initial design with four active wheels helps the robots climb onto a platform, it limits their mobility to perform precise motions. In this paper, we focus on the following challenges: 1) improve the existing hardware platform to achieve precise motion while maintaining the gap-crossing capability, 2) reduce fragility of single-connection assembly, 3) develop a planning and control algorithm to achieve a given configuration when robots do not start from aligned positions. 

\begin{figure}
    \centering
    \includegraphics[width=0.45\textwidth]{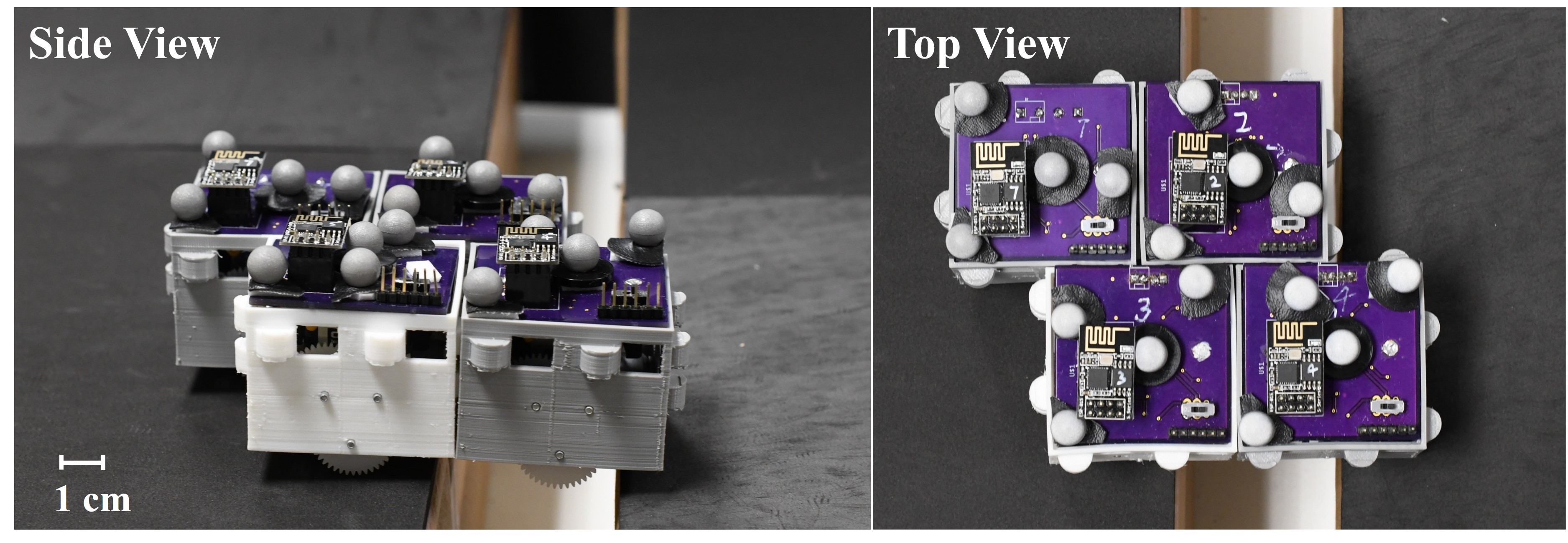}
    \caption{Four robots form a mesh configuration to cross a gap between two platforms.}
    \label{fig:cross_gap_demo_mesh}
    \vspace{-12pt}
\end{figure}

We provide solutions to the challenges mentioned above by a heterogeneous robot swarm system and a connection-pair-oriented configuration control algorithm. In this paper, we assume the target configuration is given by the user and the goal of the robots is to align and form this predefined configuration. To improve the gap-crossing performance, we introduce a heterogeneous system containing pilot robots and non-pilot robots. Pilot robots have the same design as \cite{yi2021puzzlebots} that helps climbing onto platforms. Non-pilot robots have caster wheels that enables precise control in both linear velocity and angular velocity. This heterogeneous system utilize the advantages of both designs while minimizing the drawbacks.
Based on this, we propose a \textit{connection-pair oriented configuration control algorithm} with which robots can form given configurations from unaligned positions. Due to the passive coupling mechanism, the connection between robots in \cite{yi2021puzzlebots} can be fragile and sensitive to disturbances. Borrowing the k-connectivity concept from graph theory, we introduce the mesh assembly shown in Figure~\ref{fig:cross_gap_demo_mesh}, where robots can couple in two dimensions to strengthen the connection pairs formed in one direction. Experiments show that the mesh configuration helps maintain a stable assembly formation and increases the strength of the connection. 

The outline of the paper is as follows. First, we discuss the problem setup of what the robots are expected to achieve in Section~\ref{sec:problem}. Then, in Section~\ref{sec:method}, we present our connection-pair-oriented configuration control algorithm that drives the robots to a given coupled configuration. Next, in Section~\ref{sec:hardware}, we present our heterogeneous system consisting of pilot robots and non-pilot robots. Finally, experiments of up to nine robots with line and mesh formation to cross gaps of different sizes, as well as calibration and a behavior sequence demonstration, are presented in Section~\ref{sec:exp_results}.

\section{Problem Formulation}\label{sec:problem}
Consider a set of $N$ robots on a 2D plane. Denote the poses of the robots as $\mathbf p = [p_1, p_2, \dots, p_N] \in \mathbb{R}^{3\times N}$, where each robot pose consists of its coordinates in $x$ and $y$ axis, and its heading angle, i.e. $p_i = [x_i, y_i, \theta_i]$. The control of the robots is based on unicycle model where the control input $\mathbf u \in \mathbb{R}^{2\times N}$ consists of linear velocities and angular velocities, i.e. $u_i = [v_i, \omega_i]$. The transition equation \cite{lavalle2006planning} is defined as
\begin{align}\label{eq:unicycle}
\footnotesize
\dot{p_i} = 
\begin{bmatrix}\dot{x_i}\\\dot{y_i}\\\dot{\theta_i}\end{bmatrix}
=
\begin{bmatrix}
\cos\theta_i & 0  \\
\sin\theta_i & 0 \\
0 & 1
\end{bmatrix}\begin{bmatrix}
v_i \\ \omega_i
\end{bmatrix}
= J_i \cdot u_i \text{ .}
\end{align}

\begin{figure}
\centering
\begin{subfigure}{0.14\textwidth}
\centering
\includegraphics[width=\textwidth]{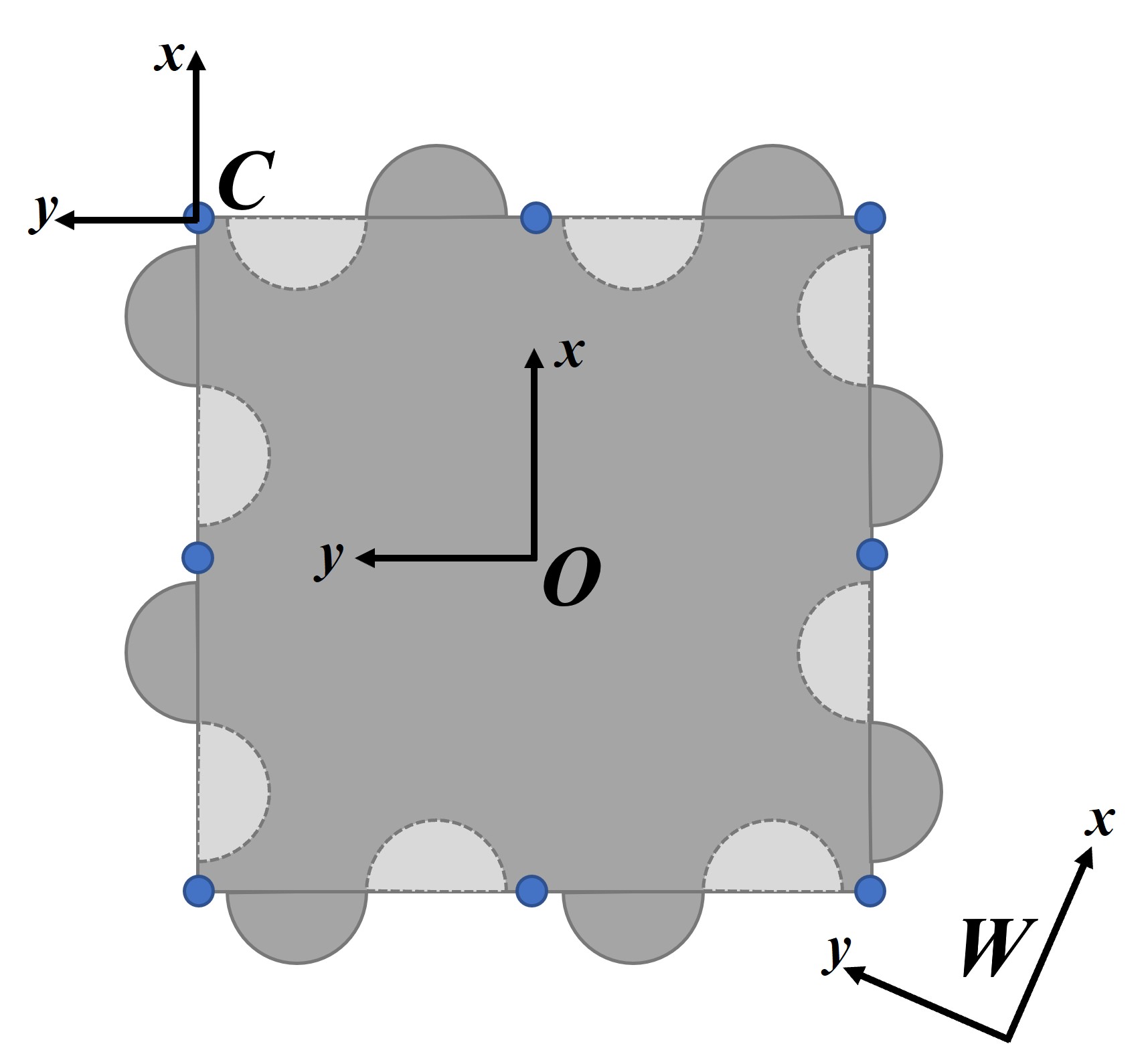}
\caption{}
\label{fig:robot_frame}
\end{subfigure}\hfill
\begin{subfigure}{0.325\textwidth}
\centering
\includegraphics[width=\textwidth]{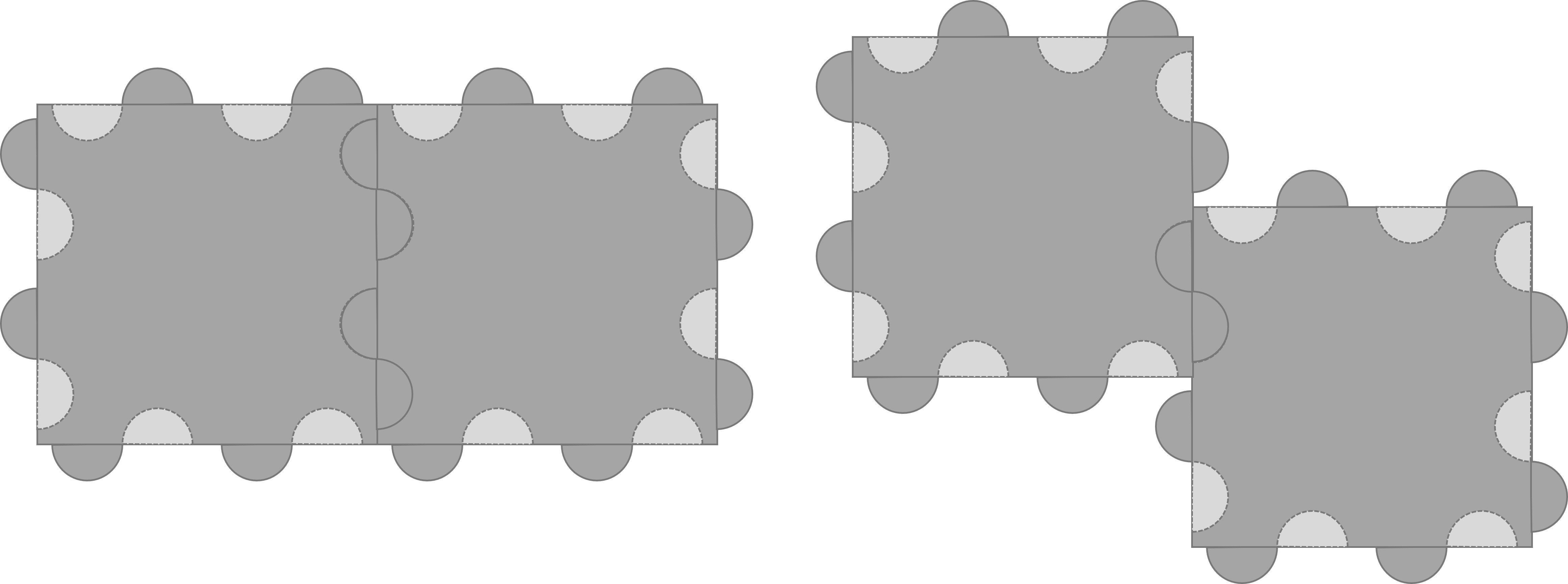}
\caption{}
\label{fig:align_robots}
\end{subfigure}
\caption{(a) PuzzleBot with eight connection points shown as blue dots and the robot frame (O), world frame (W), and connection point frame (C); (b) Two possible connection configuration for two robots.}
\label{fig:robot_alignment}
\vspace{-12pt}
\end{figure}

Our heterogeneous system consists of two type of robots - pilot and non-pilot robots. The two type of robots are different in wheel design but follow the same dynamics in Equation~\eqref{eq:unicycle}. Details will be introduced in Section~\ref{sec:hardware}. All robots have the same body with two knobs and two holes on each side to provide passive coupling behavior \cite{yi2021puzzlebots}. The knobs on one robot can be inserted into the holes of another robot to couple. To parametrize each coupling pair, we define eight connection points on the robot body as shown in Figure~\ref{fig:robot_frame}. Define the connection point set of robot $i$ as $\mathcal{C}_i$. Robot $i$ and robot $j$ are coupled when $\mathcal{C}_i \cap \mathcal{C}_j \neq \O$, forming one or more \textit{connection pairs}. Each connection pair uniquely defines a coupling configuration, while each coupling configuration may have multiple connection pairs as shown in Figure~\ref{fig:align_robots}. Define an \textit{assembly} as a group of successfully coupled robots. An assembly consists of two or more robots and the relationship of their connection pairs.

The goal configuration $\mathbf p_g = [p_{g_1}, p_{g_2}, \dots, p_{g_N}] \in \mathbb{R}^{3\times N}$ consists of $N$ relative poses on the 2D plane, i.e. $G \cdot \mathbf{p}_g^T$ is the same goal configuration as $\mathbf p_g$ where $G \in SE(2)$ is a rigid transformation on the 2D plane. To form functional assembly structures, robots in the goal configuration are coupled, i.e. for each robot $i$ in $\mathbf p_g$, there exist a robot $j$ whose $\mathcal{C}_i \cap \mathcal{C}_j \neq \O$. This coupling constraint on the goal configuration is particularly challenging since a simple go-to-goal controller cannot guarantee the coupling behavior or the alignment of the connection pairs. We will introduce our \textit{connection pair oriented configuration control} algorithm to solve this problem in section~\ref{sec:method}. 

\section{Methodology}\label{sec:method}
In this section, we will present three aspects of the configuration control based on connection pairs. Section~\ref{sec:single_pair} introduces a PID based controller that aligns a single pair of connection points. Then the formation of a two-robot assembly, which serves as a basis for multiple connection pair alignment, is defined in section~\ref{sec:problem}. Section~\ref{sec:pair_maintain} introduces an optimization-based control for an assembly to reach individual goals while maintaining in-assembly connection pairs. The last section~\ref{sec:config_control} presents our connection pair based configuration control algorithm. Heterogeneity of the system will be discussed in section~\ref{sec:hardware}.

\subsection{Single Connection Pair Alignment}\label{sec:single_pair}
As shown in Figure~\ref{fig:robot_frame}, the robot body frame is defined as $O$, and the contact frame of a connection point is defined as $C$. The $O$ frame and $C$ frame have the same orientation. Note that the $C$ frame is a general representation of a contact frame but not a specific connection point on a robot.
Given a target connection pair alignment $C_i$ and $C_j$ of the robots $i$ and $j$ respectively, where $i, j = 1, 2, \dots, N$, denote the position of $C_i$ in $x$ and $y$ axis in world frame $W$ as $[c_{x_i}, c_{y_i}]$. The heading angle of $C_i$ is $\theta_i$, aligned with the robot frame. For robot $i$, the controller for connection pair alignment is
\begin{align}\label{eq:cp_align_control}
\footnotesize
u_i = J^+_i \begin{bmatrix}
\Delta c_x \\ \Delta c_y \\ \Delta \theta
\end{bmatrix} =
J^+_i \begin{bmatrix}
c_{x_j} -  c_{x_i} \\
c_{y_j} - c_{y_i} \\
\arctan{\frac{\Delta c_y}{\Delta c_x}} - \theta_i + \theta_{bias}
\end{bmatrix} \text{,}
\end{align}
where $J^+_i$ is the pseudo-inverse of $J_i$ in Equation~\eqref{eq:unicycle}. $\Delta \theta$ is wrapped into $(-\frac{\pi}{2}, \frac{\pi}{2}]$. $\theta_{bias}$ is an angle bias added when the connection pair will lead the robots to align side-by-side. This will help the robot to push its knob into the hole of the other robot. Similarly, $u_j$ is computed accordingly as in Equation~\eqref{eq:cp_align_control}. Note that not any given set of connection pair can be aligned due to local minima with this Jacobian pseudo-inverse method. For example, for the two robots in Figure~\ref{fig:align_robots}, it will be infeasible if the connection point on the left robot is on the left side of the robot body. In our setting, we assume robots only start from feasible positions with respect to a given connection pair.

\subsection{Connection Pair Maintenance}\label{sec:pair_maintain}
Once one or more connection pairs are aligned in the system, we study the motion of an assembly - how to reach another configuration while maintaining current connection pairs within an assembly. Consider a given target control input $\mathbf{\hat u}$ for an assembly. The assembly consists of one or multiple connection pair(s) already aligned. We aim to find the control $\mathbf{u}$ that minimizes $||\mathbf{u} - \mathbf{\hat u}||^2$ while maintaining the connection pairs within the assembly. We formulate this problem as a quadratic programming (QP) problem with linear constraints. 
Define the homogeneous transformation from the world frame $W$ to the contact frame $C$ of robot $i$ to be $g_{wc}$. We have $g_{wc} = g_{wo}g_{oc}$, where $g_{oc}$ is constant for a given connection point. Denote
\begin{align*}\footnotesize
g_{oc} = \begin{bmatrix}
1&0&d_{x_c}\\
0 & 1 & d_{y_c} \\
0 & 0 & 1
\end{bmatrix} \text{,}
\end{align*}
and when applying $u=[v_i, \omega_i]$ over time $\Delta t$,
\begin{align*}\footnotesize
g_{wo} =\begin{bmatrix}
\cos(\theta_i+\omega_i \Delta t) & -\sin(\theta_i+\omega_i \Delta t) & x_i+v_i \Delta t\cos\theta_i \\
\sin(\theta_i+\omega_i \Delta t) & \cos(\theta_i+\omega_i \Delta t) & y_i+v_i \Delta t\sin\theta_i \\
0 & 0 & 1
\end{bmatrix} \text{ .}
\end{align*}
By calculating $g_{wc} = g_{wo}g_{oc}$ and extracting the translational component of $C$, the position vector $[c_{x_i}, c_{y_i}]^T$ becomes
\begin{align*} \footnotesize
\begin{bmatrix}
x_i+v_i \Delta t\cos\theta_i  + d_{x_c}\cos(\theta_i+\omega_i \Delta t) - d_{y_c}\sin(\theta_i+\omega_i \Delta t)\\
y_i+v_i \Delta t\sin\theta_i + d_{x_c}\sin(\theta_i+\omega_i \Delta t) + d_{y_c}\cos(\theta_i+\omega_i \Delta t)
\end{bmatrix} \text{ .}
\end{align*}
Consider $\Delta t$ as a very small value, we may perform Taylor expansion around $\theta_i$ to linearize the above equation as
\begin{align}
\footnotesize
\cos(\theta_i+\omega_i \Delta t) &\approx
\cos\theta_i - \omega_i \Delta t\sin\theta_i \\
\sin(\theta_i+\omega_i \Delta t) &\approx
\sin\theta_i + \omega_i \Delta t\cos\theta_i  \text{ .}
\end{align}
To simplify the notation, denote $\cos\theta_i$ as $c_i$ and $\sin\theta_i$ as $s_i$, the position vector becomes
\begin{align} \label{eq:linear_con} \footnotesize
\begin{bmatrix}
c_{x_i} \\ c_{y_i} 
\end{bmatrix} =\begin{bmatrix}
c_i & -d_{x_c}s_i-d_{y_c} c_i \\
s_i & d_{x_c} c_i-d_{y_c} s_i
\end{bmatrix} 
u_i\Delta t +
\begin{bmatrix}
x_i + d_{x_c} c_i - d_{y_c} s_i\\
y_i+ d_{x_c} s_i+ d_{y_c} c_i
\end{bmatrix} \text{ .}
\end{align}
The connection pair constraint on robot $i$ and $j$ is defined as
\begin{align}\label{eq:cp_constraint}
-\epsilon \leq \begin{bmatrix}
c_{x_i} \\ c_{y_i} 
\end{bmatrix} - \begin{bmatrix}
c_{x_j} \\ c_{y_j} 
\end{bmatrix} \leq \epsilon   \text{ .}
\end{align}
We may stack all the connection pair constraints in Equation~\eqref{eq:cp_constraint} in an assembly in one equation $A \mathbf u \leq b$. The connection pair maintenance problem then becomes
\begin{align}\label{eq:cp_opt}
\mathbf{u}^{*} &= \argmin ||\mathbf{u} - \mathbf{\hat u}||^2 \\
& \text{s.t. }A \mathbf u \leq b \text{ .}
\end{align}

\subsection{Connection Pair Based Configuration Control}\label{sec:config_control}

\begin{figure}
\centering
\begin{subfigure}{0.16\textwidth}
\includegraphics[width=\textwidth]{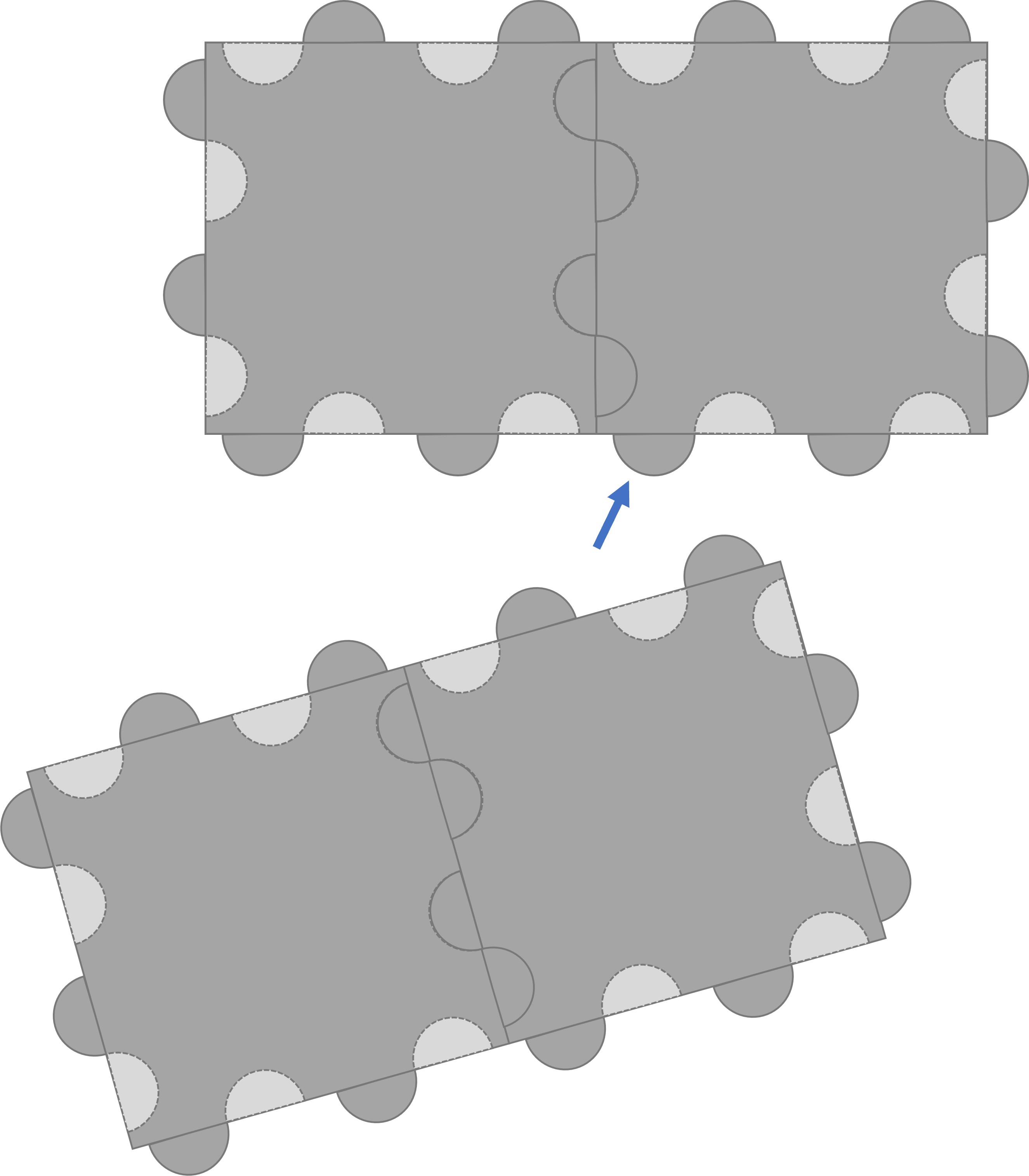}
\caption{}\label{fig:form_convex}
\end{subfigure}
\begin{subfigure}{0.19\textwidth}
\includegraphics[width=\textwidth]{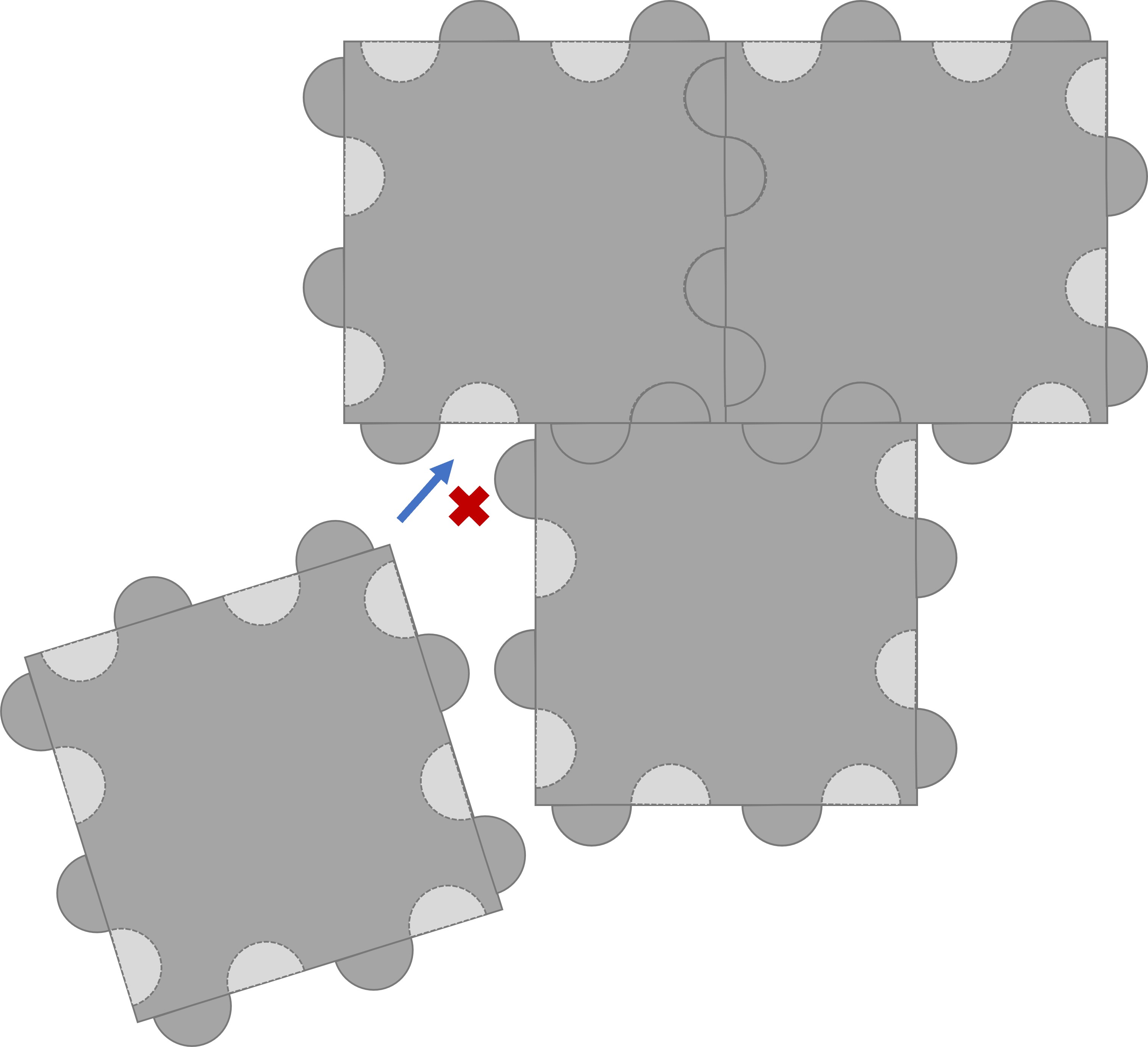}
\caption{}\label{fig:form_concave}
\end{subfigure}
\caption{(a) Two convex assemblies couple to form a larger assembly. (b) One robot cannot couple with a concave assembly.}
\label{fig:form_convex_concave}
\vspace{-12pt}
\end{figure}

As defined in section~\ref{sec:problem}, given a goal configuration $\mathbf p_g$, we aim to provide a solution for the robots to reach this goal configuration. During the process of assembling into the goal configuration, a sequential constraint exists - concave assemblies may not be able to assemble into one larger assembly due to the knobs blocking each other. For example, in Figure~\ref{fig:form_convex}, two convex assemblies, each formed by two robots, are able to couple and form a larger assembly of four robots. However, in Figure~\ref{fig:form_concave}, one robot tries to couple with a concave assembly of three robots, but the knobs block its motion. In the algorithm to be introduced in Section~\ref{sec:config_control}, we plan to avoid forming concave assemblies by prioritizing convex assemblies first.

\begin{algorithm}[h]
\caption{Find Existing Connection Pairs}
\label{alg:exist_cps}
    \begin{algorithmic}[1]
    \Input{$\mathbf p$: input poses}
    \Output{$pair\_dict$: connection pairs}
    \Initialize{$pair\_dict$=\{\}}
    \Function{findExistPairs}{$\mathbf p$}
    \State $\mathcal{G}$ $\gets$ constructGraph($\mathbf p$) based on distance
    \State $edge\_set$ $\gets$ getMinimumSpanningTree($\mathcal{G}$)
    \For{vertex $i$ and $j$ in $edge\_set$}
    \State [$C_i$, $C_j$] $\gets$ findMinDistancePair($\mathcal{C}_i$, $\mathcal{C}_j$)
    \State $pair\_dict$[($i$, $j$)] = [$C_i$, $C_j$]
    \EndFor
    \State \Return{$pair\_dict$}
    \EndFunction
    \end{algorithmic}
\end{algorithm}

The connection pair assignment is shown in Algorithm~\ref{alg:cp_generate}. First, we align the center of the input goal configuration with the current robot poses as
\begin{align}
\mathbf p_g' = \mathbf p_g - \overline{\mathbf p_g} + \overline{\mathbf p}  \text{,}
\end{align}
where $\overline{\mathbf p}$ is the mean of $\mathbf p$. We then find the connection pair assignment based on the goal configuration $\mathbf{p}_g'$ as shown in Algorithm~\ref{alg:exist_cps}. In Algorithm~\ref{alg:exist_cps}, we find the connection pairs based on an existing configuration. This input robot poses $\mathbf p$ assumes the robots are already aligned. First, we construct a fully connected graph $\mathcal{G}$ where the vertices are the location points in $\mathbf p$ and the edge weights are the distances between the vertices. We then find the Minimum Spanning Tree (MST) based on this distance-induced graph. This provides information on the minimum connection pairs to monitor in order to maintain the current configuration $\mathbf p$. Then for each edge, we loop through all combinations of connection points on the two vertices of this edge to find the one with minimum distance. With this information in Algorithm~\ref{alg:cp_generate}, we obtain the connection pairs needed to form the goal configuration $\mathbf p_g$. The pairs are then sorted based on the distance between the robot poses in $\mathbf p_g$. Notice that the concave assembly is only formed with mesh configuration, and line formation can only form a convex assembly. Two robots in mesh configuration always have a distance of $\sqrt{2}$ times the body length between each other, which is larger than that of the distance in the line configuration. By sorting the distance between robot poses, we can guarantee that convex assemblies are always formed before the concave assemblies. We then calculate the distance matrix between $\mathbf p_g'$ and $\mathbf p$. Each element in the distance matrix $d_{ij}$ is calculated as $d_{ij} = ||p_{g_i}' - p_j||^2$. Hungarian algorithm \cite{kuhn1955hungarian} is then used to find the minimum sum of distance assignment between the shifted goal configuration and the current robot poses. Finally, the resulting assignment is mapped to the sorted goal configuration pairs.

\begin{algorithm}[h]
\caption{Connection Pair Assignment}
\label{alg:cp_generate}
    \begin{algorithmic}[1]
    \Input{$\mathbf p_g$: goal configuration, $\mathbf p$: robot poses}
    \Output{$pair\_dict$: connection pair assignment}
    \Function{assignConnectionPairs}{$\mathbf p_g$, $\mathbf p$}
    \State $\mathbf{p}_g'$ $\gets$ alignCenter($\mathbf{p}_g$,  $\mathbf{p}$)
    \State $goal\_pair\_dict$ $\gets$ findExistPairs($\mathbf{p}_g'$)
    \State $sorted\_pair\_dict$ $\gets$ sortPairs($goal\_pair\_dict$, $\mathbf{p}_g'$) based on distance
    \State $dist\_matrix$ $\gets$ getDistanceMatrix($\mathbf{p}_g'$, $\mathbf{p}$)
    \State $id\_assign$ $\gets$ HungarianAlgorithm($dist\_matrix$)
    \State $pair\_dict$ $\gets$ updatePairAssignment($sorted\_pair\_dict$, $id\_assign$)
    \State \Return{$pair\_dict$}
    \EndFunction
    \end{algorithmic}
\end{algorithm}

The configuration control algorithm is shown in Algorithm~\ref{alg:config_control}. It takes in a goal configuration and outputs the control input for the robots to execute. During the process, we maintain the information of 1) busy vector where $busy[i]$ represents if robot $i$ is currently aligning an active connection pair, 2) already connected pairs, denoted as $\mathcal{C}_{conn} = \{(i, j): (c_i, c_j), \dots| c_i\in \mathcal{C}_i, c_j\in \mathcal{C}_j, \mathcal{C}_i \cap \mathcal{C}_j = (c_i, c_j)\}$; 3) the active connection pairs $\mathcal{C}_{exec}$ that are currently being executed. First, we obtain the goal connection pairs to be executed, denoted as $\mathcal{C}_{goal}$ with Algorithm~\ref{alg:cp_generate}. To maintain a dynamic connection between the robots, we will updated the connection pairs in $\mathcal{C}_{conn}$ with Algorithm~\ref{alg:exist_cps}. Then we check if any new connection pair can be executed, as in the single pair alignment in Section~\ref{sec:single_pair} and the target control input obtained in this step is denoted as $\mathbf{\hat u}[busy]$. Note that, only the robots that have active connection pairs to execute are assigned with target control input. Therefore, for already connected pairs, e.g. $(i,j) \in \mathcal{C}_{conn}$, if robot $i$ is $busy$, the target control input for robot $j$ becomes $\hat u_j = \hat u_i$ to mimic the motion of the leading robot $i$ in this connection pair. To limit the influence of uncertainties in the hardware actuation, we incorporate a connection bias in the control signal to further maintain the already connected pairs. For each robot $i$, the connection bias is
\begin{align*}
u_{connection\_bias}[i] = \sum_{j}J^{+}_i(p_j - p_i), \text{for all} (i,j) \in \mathcal{C}_{conn}  \text{ .}
\end{align*}
The optimal control input $\mathbf{u}^*$ is then obtained from Equation~\eqref{eq:cp_opt}. Finally, we find the already aligned connection pairs in the current execution set $\mathcal{C}_{exec}$. The connected pairs are removed, and the robots return to non-busy status. Depending on the structure of the configuration, the best case run time of this algorithm is $O(\log N)$ while the worst case is $O(N)$.

\begin{algorithm}[h]
\caption{Connection Pair Based Configuration Control}
\label{alg:config_control}
    \begin{algorithmic}[1]
    \Input{$\mathbf p_g$: goal configuration, $\mathbf p$: robot poses, $\epsilon$: threshold}
    \Output{$\mathbf u^*$: control input}
    \Initialize{$busy =$[False, \dots], $\mathcal{C}_{conn}=\{\}$, $\mathcal{C}_{exec}=\{\}$}
    \Function{ConfigurationControl}{$\mathbf p_g$, $\mathbf p$}
    \State $\mathcal{C}_{goal}$ $\gets$ assignConnectionPairs($\mathbf p_g$, $\mathbf p$)
    \For{$pairs$ in $\mathcal{C}_{conn}$}
    \State update connection pairs $pairs$
    \EndFor
    \For{($i$, $j$) in $\mathcal{C}_{goal}$}
    \If{$i$, $j$ not busy}
    \State $\mathcal{C}_{exec}$.append(($i$, $j$))
    \State $busy$[$i$] = True, $busy$[$j$] = True
    \EndIf
    \EndFor
    \State $\mathbf{\hat u}[busy]$ $\gets$ alignConnectionPairs($\mathcal{C}_{exec}$, $\mathbf p$)
    \State $\mathbf{\hat u}[\sim busy]$ $\gets$ $\mathbf{\hat u}[busy]$ for each $\mathcal{C}_{conn}$
    \State $\mathbf{\hat u} = (1-k_{bias})\mathbf{\hat u} + k_{bias} \mathbf{u}_{connection\_bias}$
    \State $\mathbf{u}^* = \argmin ||\mathbf{u} - \mathbf{\hat u}||^2$, $\text{s.t. }A \mathbf u \leq b$
    \For{($i$, $j$) in $\mathcal{C}_{exec}$}
    \If{distance between connection pairs $<$ $\epsilon$}
    \State $\mathcal{C}_{exec}$.remove(($i$, $j$))
    \State $busy$[$i$] = False, $busy$[$j$] = False
    \State $u_i$, $u_j$ $\gets$ $0$
    \EndIf
    \EndFor
    \State \Return{$\mathbf u^*$}
    \EndFunction
    \end{algorithmic}
\end{algorithm}

\section{Hardware Setup}\label{sec:hardware}
Each PuzzleBot is equipped with onboard power, computation, communication, and actuation. The circuit design remains the same as in the previous version of the PuzzleBots \cite{yi2021puzzlebots}. Each robot is equipped with four trackers for indoor localization via the Vicon motion tracking system. The robots, Vicon, and a central computer are connected within the same WiFi network. The central computer computes the command velocity and sends it to each robot to execute.

\begin{figure}
\centering
\begin{subfigure}{0.115\textwidth}
\centering
\includegraphics[width=\textwidth]{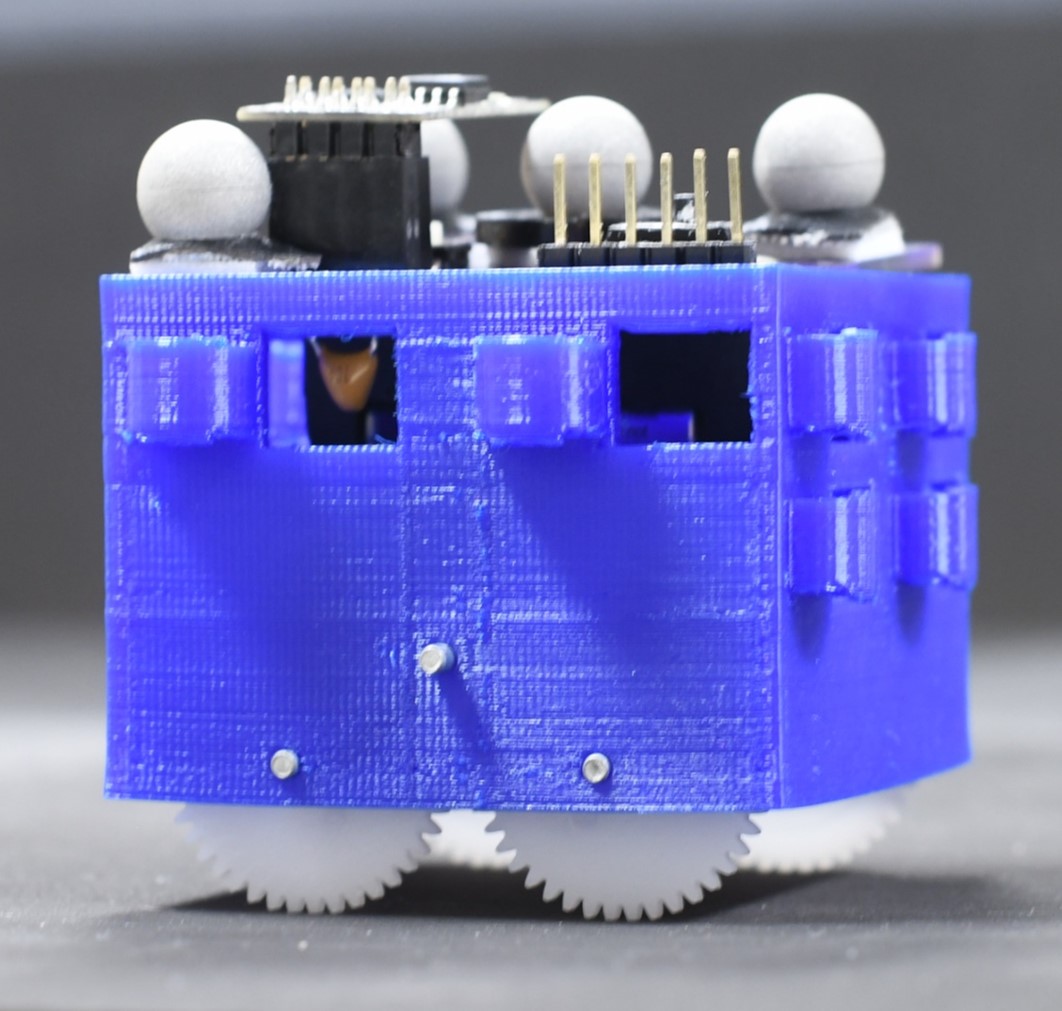}
\caption{}\label{fig:pilot_side}
\end{subfigure}
\begin{subfigure}{0.115\textwidth}
\centering
\includegraphics[width=\textwidth]{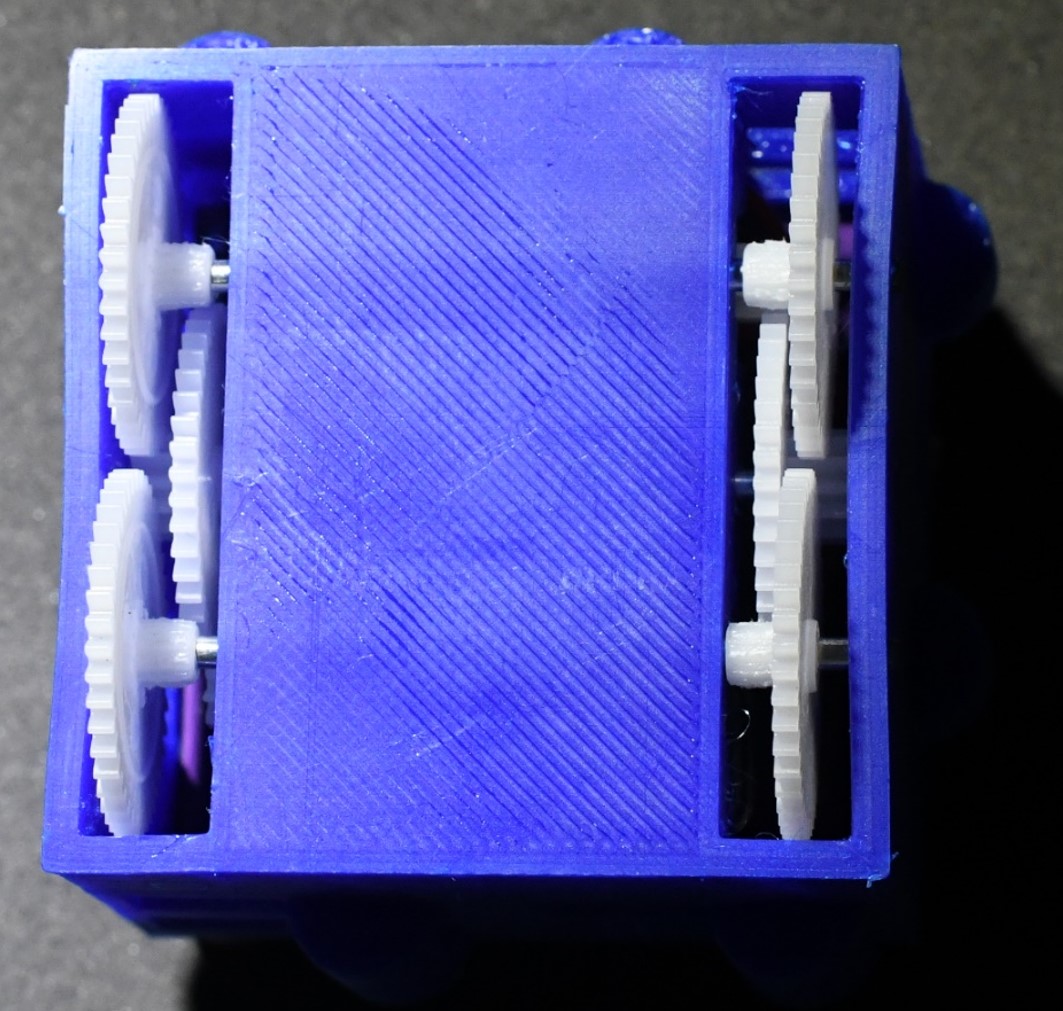}
\caption{}\label{fig:pilot_bottom}
\end{subfigure}
\begin{subfigure}{0.115\textwidth}
\centering
\includegraphics[width=\textwidth]{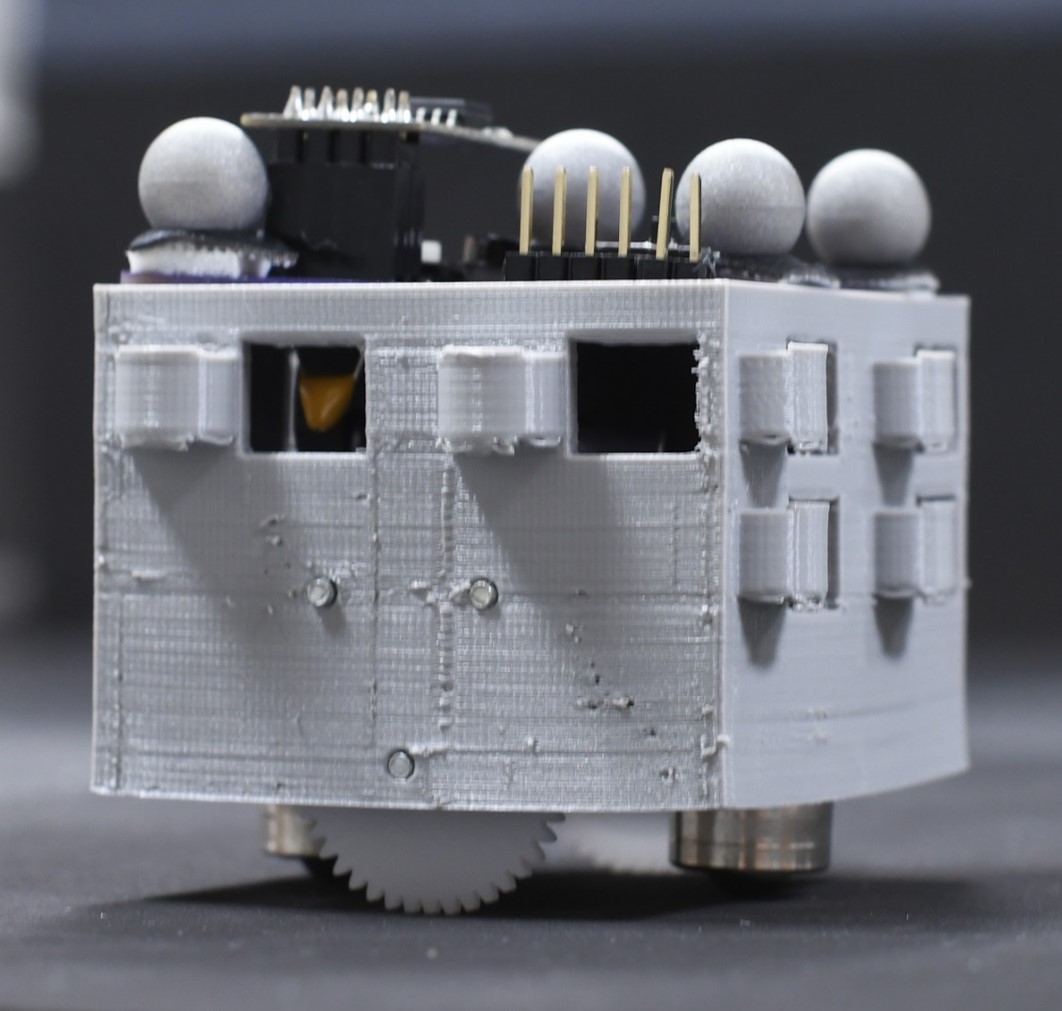}
\caption{}\label{fig:nonpilot_side}
\end{subfigure}
\begin{subfigure}{0.115\textwidth}
\centering
\includegraphics[width=\textwidth]{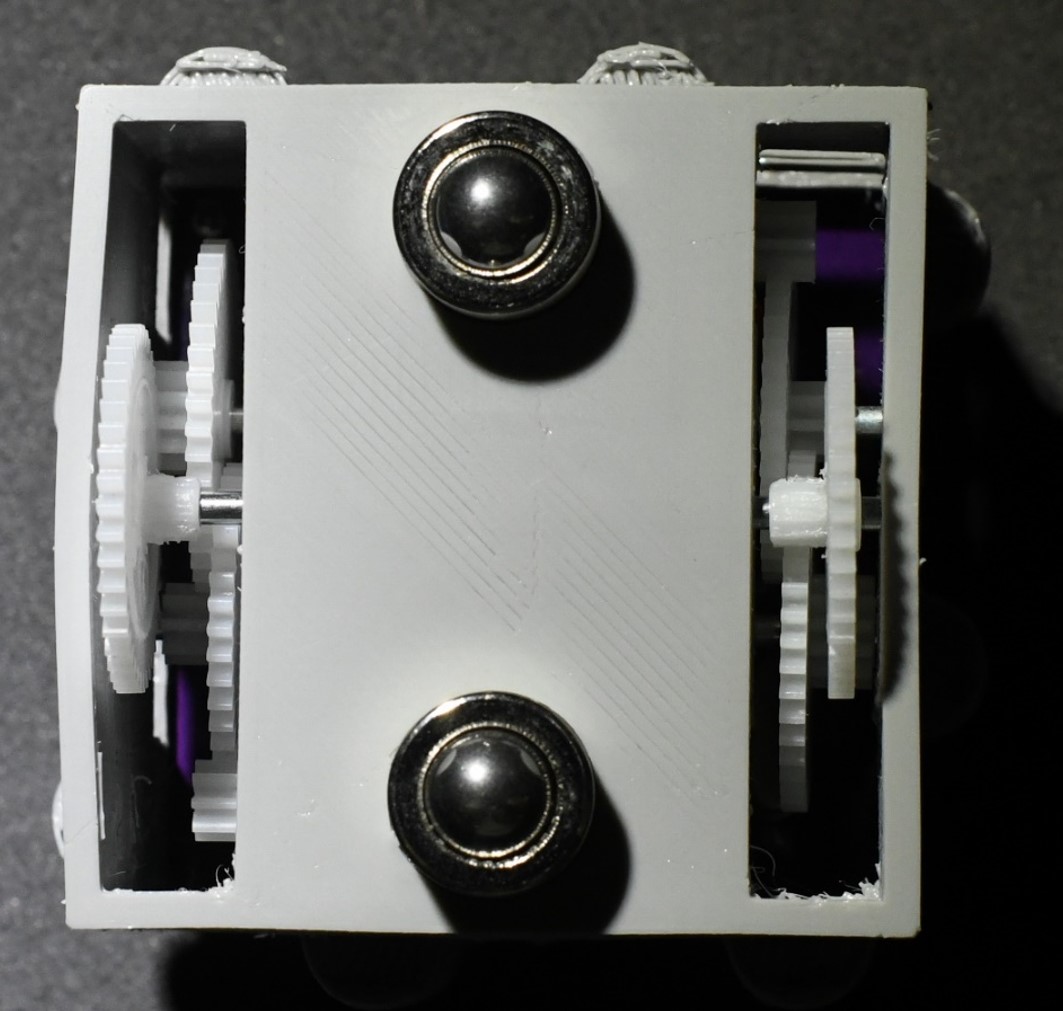}
\caption{}\label{fig:nonpilot_bottom}
\end{subfigure}
\caption{Pilot robot: (a) side view, (b) bottom view. Non-pilot robot: (c) side view, (d) bottom view.}
\label{fig:pilot_nonpilot}
\vspace{-12pt}
\end{figure}

In the previous version of the PuzzleBots, each robot has two wheels on each side, four wheels in total. The two wheels are controlled by one motor via a double reduction gear set. This four-wheeled setup successfully demonstrated the ability to climb onto a platform when crossing a gap. However, their mobility is limited due to friction. The robot can freely move forward and backward but has limited rotation ability. This highly restricts the robots' performance of achieving precise poses. Therefore, we modified the design by adding two caster wheels in the front and back while replacing the two wheels on each side with only one. The new robot design is shown in Figure~\ref{fig:nonpilot_side}. While the caster wheels successfully solve the problem in rotation and enable the robot to do high-precision tasks, the gap-crossing behavior becomes limited. Since the caster wheels are not actuated, the robot cannot climb onto the platform when the caster wheel reaches the other side of the gap. Thus, we propose a heterogeneous robot swarm system that consists of two types of robots: pilot robots and non-pilot robots. The pilot robot is a four-wheeled robot, and the non-pilot robot is the one with casters and side wheels. As shown in Figure~\ref{fig:pilot_nonpilot}, the wheels of the pilot robot will help to climb onto a platform, while the flexibility of the non-pilot robot enables the system to form complex configurations and perform high-precision tasks.

The electronics board on the pilot and non-pilot robot is the same. The control of these two kinds of robots also remains the same, i.e., both follow the differential drive model. The linear velocity is linearly proportional to the left and weight wheel average velocity. The angular velocity is linearly proportional to the velocity difference between the right and left wheels. Denote the left and right motor Pulse-width modulation (PWM) signal as $M_r$, $M_l$. The motor rotational speed is linearly proportional to the PWM signal with a fixed load. We parametrize the velocity equation for each robot as
\begin{align}\label{eq:motor_eq}
v = k_v \frac{M_r + M_l}{2}, 
\omega = k_{\omega}(M_r - M_l) \text{ .}
\end{align}
From experiments, each robot needs a start-up torque to move. Denote the corresponding PWM signal as $M_{min}$, and a maximum PWM as $M_{max}$. With a given control input $u^*=[v^*, \omega^*]$, we compute
\begin{align*}{\scriptstyle
\argmin_{M_r, M_l} \mu_v || k_v \frac{M_r + M_l}{2} - v^*||^2 + \mu_{\omega} || k_{\omega} (M_r - M_l) - \omega^*||^2} \\
M_{min} \leq |M_r, M_l| \leq M_{max} \text{,}
\end{align*}
where $\mu_v$ and $\mu_{\omega}$ are weight parameters for linear and angular velocities respectively.
The difference of controlling the pilot and the non-pilot robot lies in their parameters of the control feasibility region, which is the constraints $A \mathbf{u} \leq b$ in Equation~\eqref{eq:cp_opt}. We will present the calibration of the feasibility region in Section~\ref{sec:calibration} as well.

\section{Experiments and Results}\label{sec:exp_results}
The system consists of several PuzzleBots, a Vicon localization system, and a central computer within the same network. The algorithm is first tested in simulation in CoppeliaSim\cite{coppeliaSim} with the Vortex Studio\footnote{\url{https://www.cm-labs.com/vortex-studio/}} as the physics engine for concave objects. We conducted three sets of experiments: Section~\ref{sec:calibration} presents the hardware calibration of the pilot and non-pilot robots. Section~\ref{sec:combined_seq} shows a set of screenshots from a video sequence where three non-pilot robots and one pilot robot couple into a mesh configuration, cross a gap and decouple on the other platform. The last Section~\ref{sec:gap_cross} presents quantitative results of the gap-crossing performance based on the line and mesh configuration.

\subsection{Robot Calibration}\label{sec:calibration}

\begin{figure}
\centering
\begin{subfigure}{0.235\textwidth}
\centering
\includegraphics[width=\textwidth]{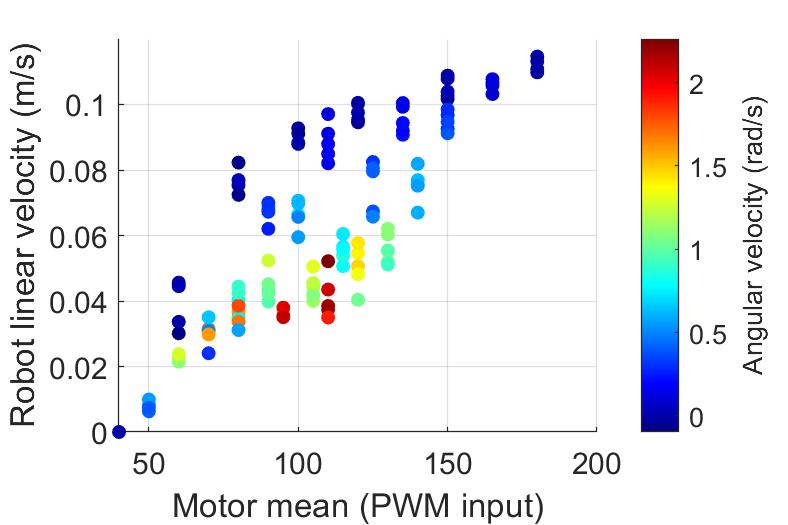}
\caption{Non-pilot robot linear velocity}\label{fig:calib_v_nonpilot_color}
\end{subfigure}
\begin{subfigure}{0.235\textwidth}
\centering
\includegraphics[width=\textwidth]{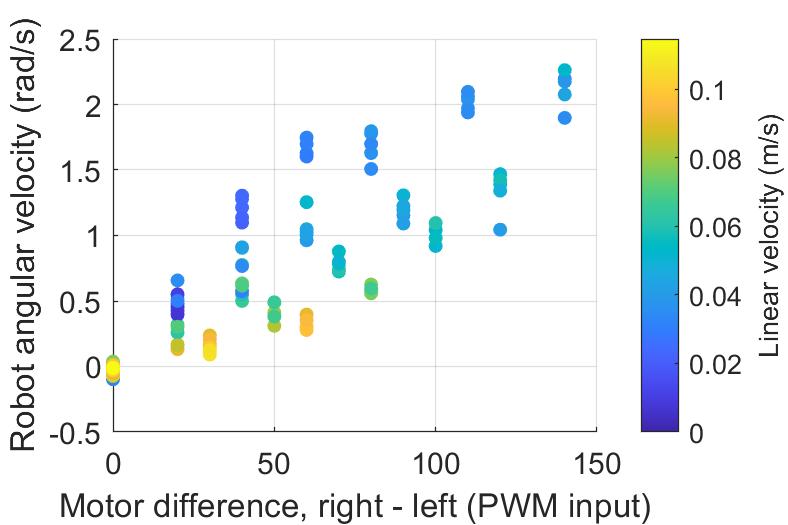}
\caption{Non-pilot robot angular velocity}\label{fig:calib_w_nonpilot_color}
\end{subfigure}
\begin{subfigure}{0.235\textwidth}
\centering
\includegraphics[width=\textwidth]{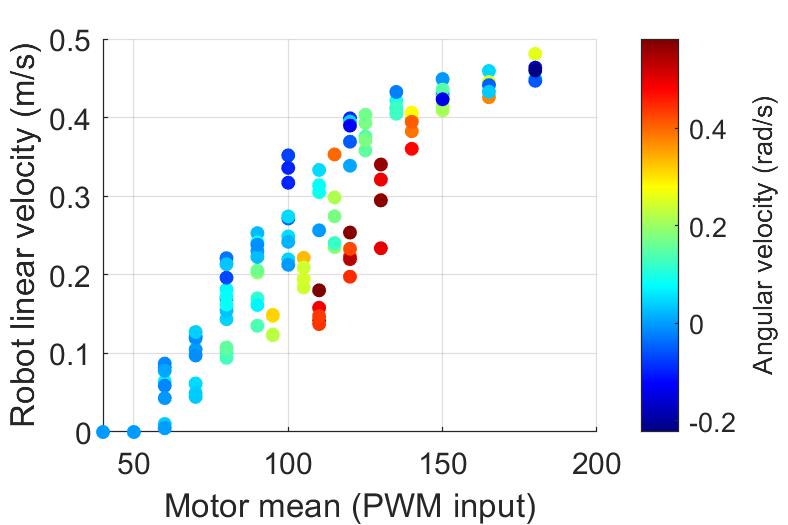}
\caption{Pilot robot linear velocity}\label{fig:calib_v_pilot_color}
\end{subfigure}
\begin{subfigure}{0.235\textwidth}
\centering
\includegraphics[width=\textwidth]{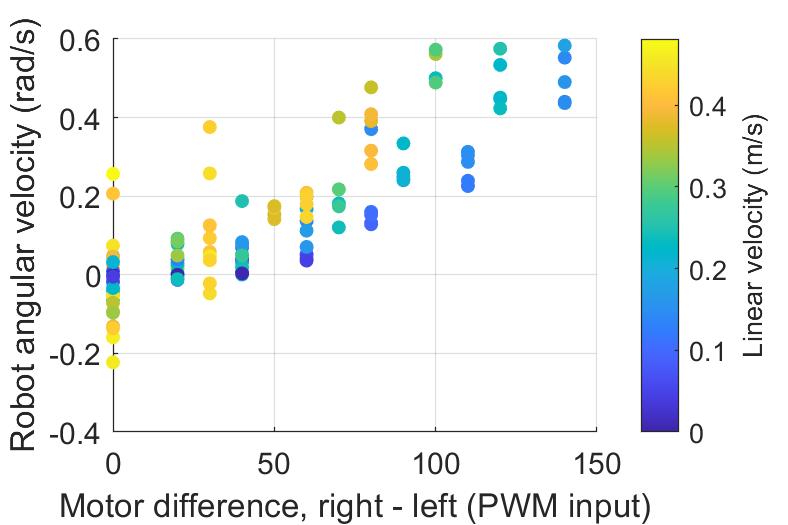}
\caption{Pilot robot angular velocity}\label{fig:calib_w_pilot_color}
\end{subfigure}
\begin{subfigure}{0.235\textwidth}
\centering
\includegraphics[width=\textwidth]{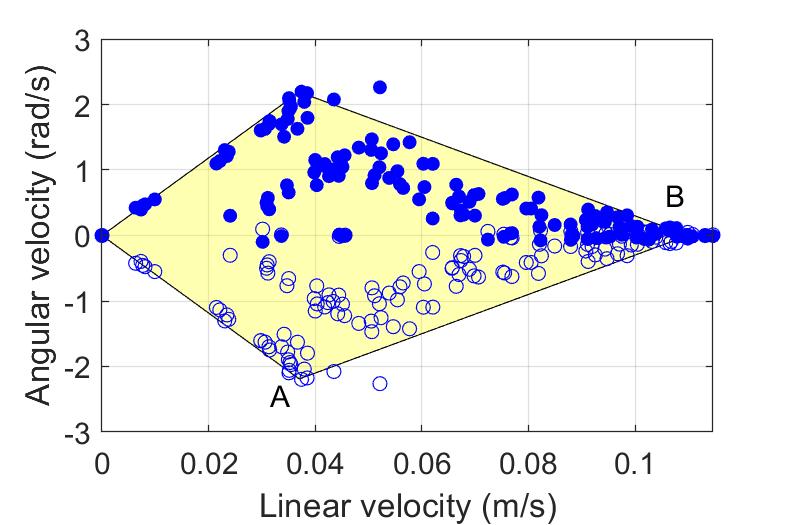}
\caption{Non-pilot robot $v-\omega$ constraints}\label{fig:calib_vw_nonpilot_color_polyg}
\end{subfigure}
\begin{subfigure}{0.235\textwidth}
\centering
\includegraphics[width=\textwidth]{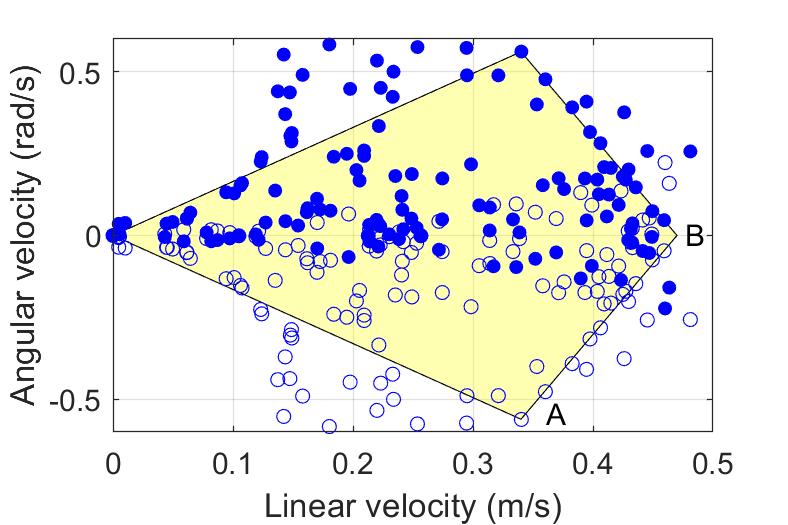}
\caption{Pilot robot  $v-\omega$ constraints}\label{fig:calib_vw_pilot_color_polyg}
\end{subfigure}
\caption{Calibration results of non-pilot and pilot robot. (a) - (d): linear and angular velocities with respect to PWM input (0 to 255). (e) (f): feasibility region in terms of $v-\omega$ of the non-pilot and pilot robots.}
\label{fig:calibration}
\vspace{-15pt}
\end{figure}

Robots are commanded with a combination of different motor PWM signals for a period of time, and the Vicon software records the poses. Due to the noise, the velocity obtained directly from two consecutive poses is unusable. Thus, we average the linear and angular velocity across a longer period of time (several seconds). We calculated the mean and difference of the left and right PWM signals, and the result is shown in Figure~\ref{fig:calibration}. The parameters obtained from the experiments are $k_{v,\text{non-pilot}}=0.0006$, $k_{\omega,\text{non-pilot}}=0.0142$, $k_{v,\text{pilot}}=0.0024$, $k_{\omega,\text{pilot}}=0.0033$. Figure~\ref{fig:calib_v_nonpilot_color} and \ref{fig:calib_v_pilot_color} show that with the same motor mean, the robot has higher linear velocity when the angular velocity is small. As shown in Figure~\ref{fig:calib_vw_nonpilot_color_polyg} and \ref{fig:calib_vw_pilot_color_polyg}, the feasibility region of the non-pilot and pilot when the linear velocity $v$ is positive is shown in the polygon. The points obtained from the experiment are projected along the $\omega=0$ axis based on symmetry. Similarly, the polygon is projected onto the negative $v$ plane. The point $A=(a_x, a_y)$ and point $B=(b_x, 0)$ are the critical points defining the polygon. We manually label $A$ and $B$ by observation, and the polygon region defined by $A$ and $B$ is recorded.
In our experiment, we have $A_{pilot}=(0.34,-0.561)$, $B_{pilot}=(0.47,0)$, $A_{non-pilot}=(0.037,-2.19)$, $B_{pilot}=(0.11,0)$. The non-pilot robot is able to rotate in high angular velocity with low linear velocity, while the pilot robot requires a large linear velocity to rotate. The steering distance for the pilot robot to turn is thus larger, making it difficult to perform the coupling behavior, which requires precise rotation. Therefore, a combination of both robots can utilize the flexibility of the non-pilot robot, as well as the climbing wheels of the pilot robot.

\subsection{Combined Behavior Sequences}\label{sec:combined_seq}

\begin{figure*}
\centering
\includegraphics[width=\textwidth]{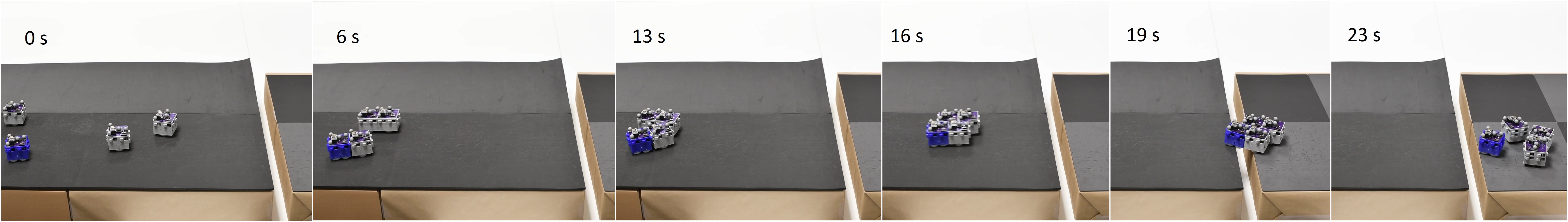}
\caption{Screenshots of four robots coupling to form a mesh configuration, crossing the gap, and decouple.} \label{fig:cross_screenshots}
\vspace{-5pt}
\end{figure*}

As shown in Figure~\ref{fig:cross_screenshots}, the four robots - three non-pilot robots shown in grey and one pilot robot shown in blue, start from unaligned positions on the left platform. The two platforms have a height difference of 5 mm, and the gap size is 40 mm. Due to the large steering distance of pilot robots, they are not given a velocity command until they are coupled with non-pilot robots, which is embedded in the algorithm. In this case, we see that the robots are able to first form two line assembly based on the connection pair assignment and then come together to form a mesh configuration. They are able to cross the gap and reach the other platform, and eventually decouple with each other.

\subsection{Gap-crossing Performances}\label{sec:gap_cross}

\begin{figure}
\centering
\begin{subfigure}{0.235\textwidth}
\centering
\includegraphics[width=\textwidth]{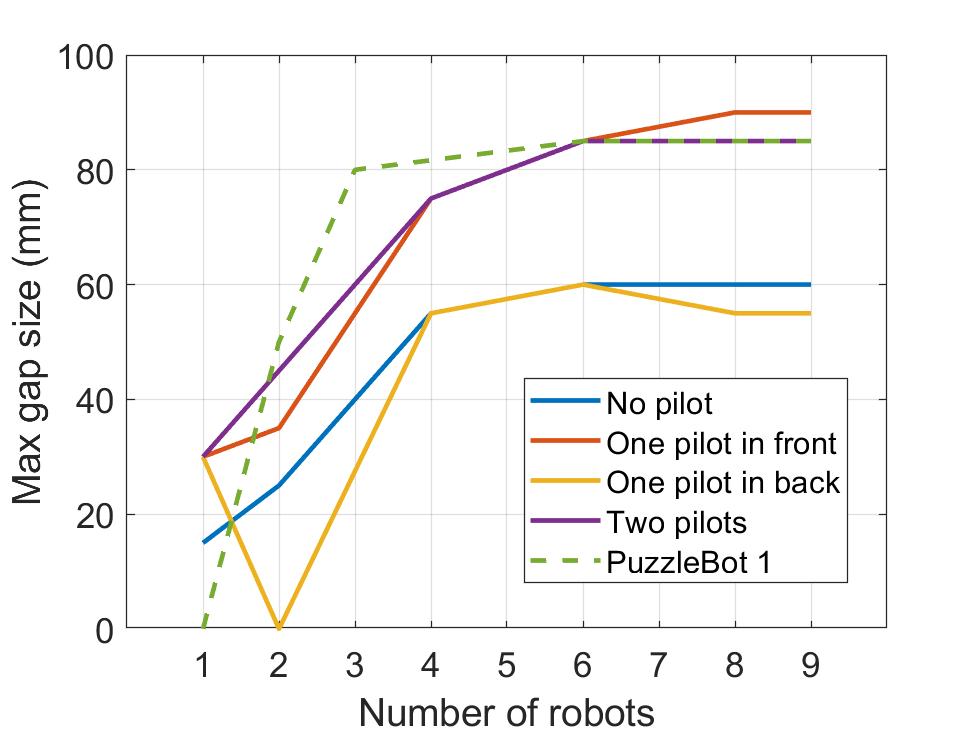}
\caption{Line configuration}\label{fig:max_gap_line}
\end{subfigure}
\begin{subfigure}{0.235\textwidth}
\centering
\includegraphics[width=\textwidth]{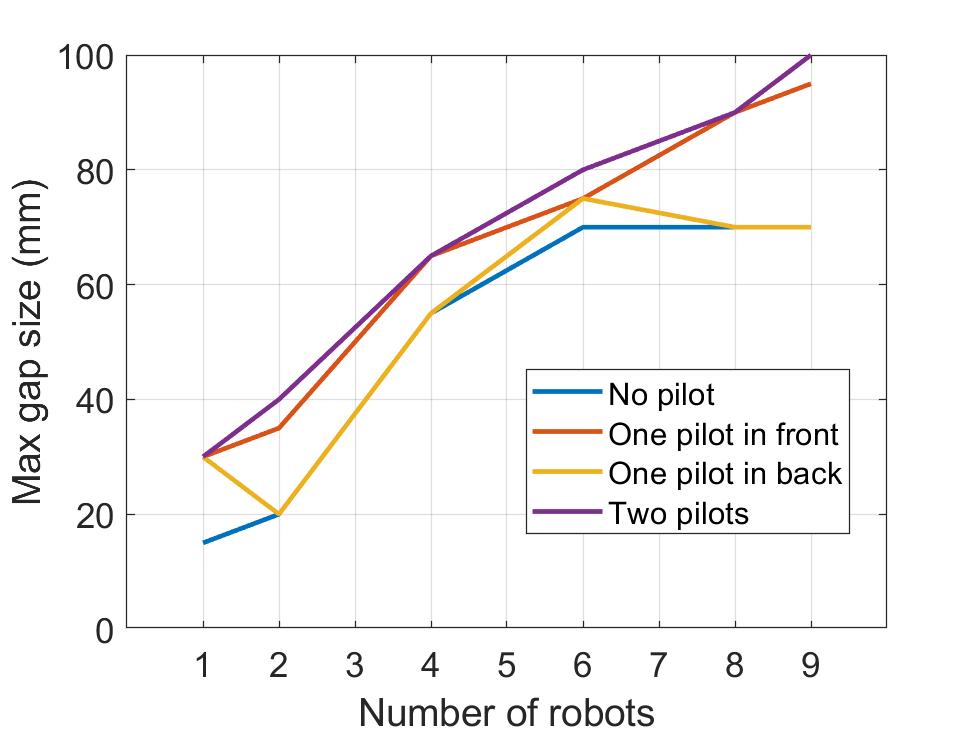}
\caption{Mesh configuration}\label{fig:max_gap_mesh}
\end{subfigure}
\begin{subfigure}{0.235\textwidth}
\centering
\includegraphics[width=\textwidth]{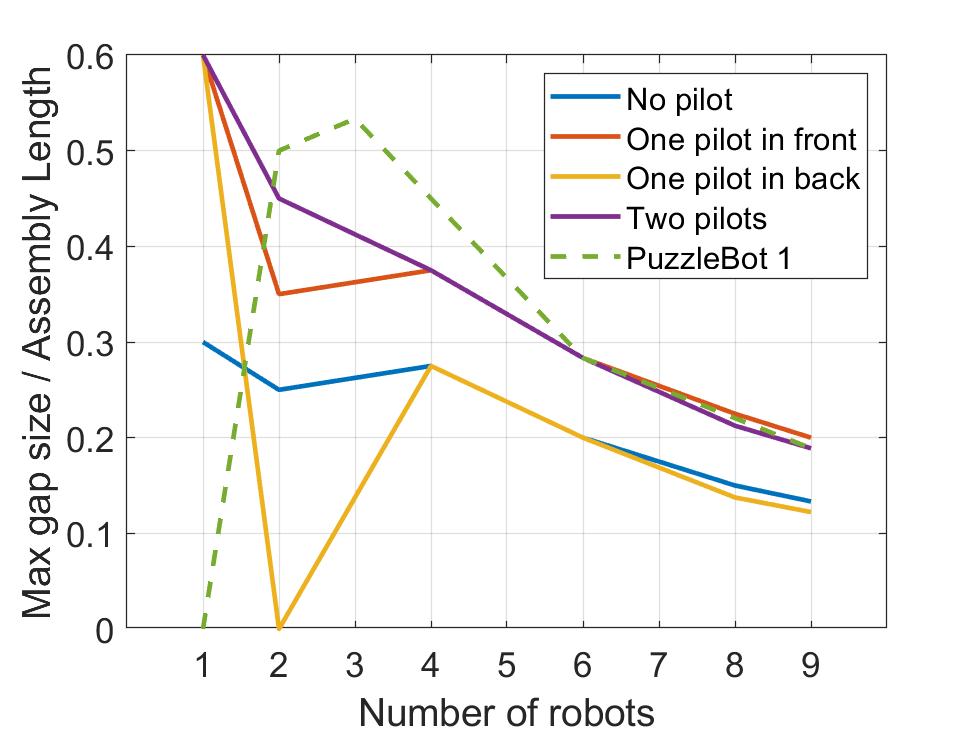}
\caption{Line configuration}\label{fig:assembly_length_line}
\end{subfigure}
\begin{subfigure}{0.235\textwidth}
\centering
\includegraphics[width=\textwidth]{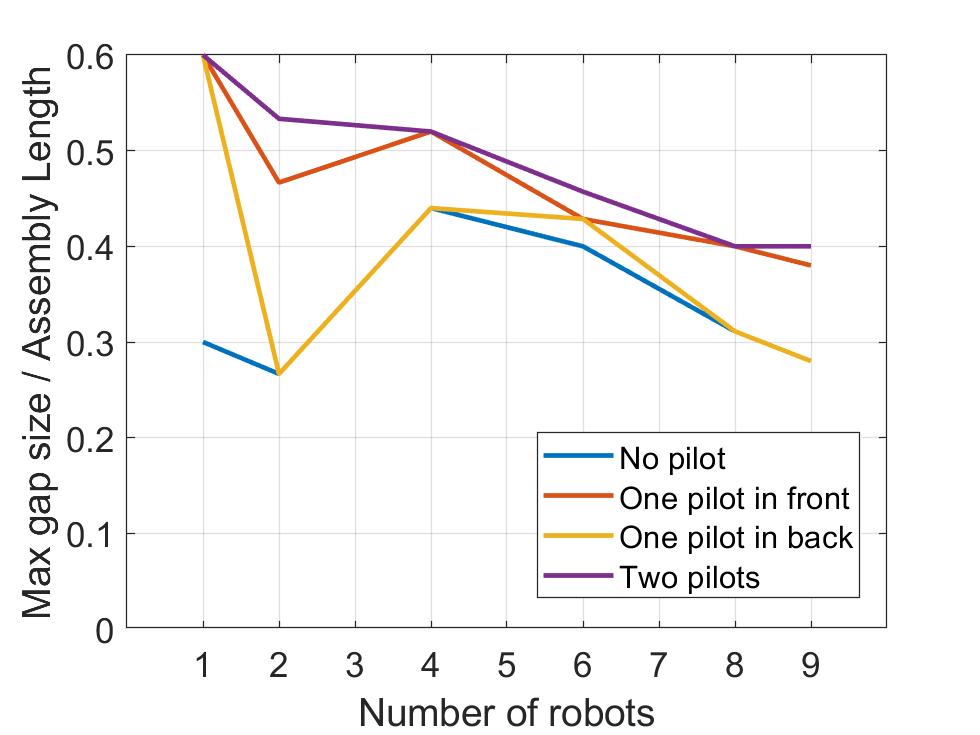}
\caption{Mesh configuration}\label{fig:assembly_length_mesh}
\end{subfigure}
\begin{subfigure}{0.235\textwidth}
\centering
\includegraphics[width=\textwidth]{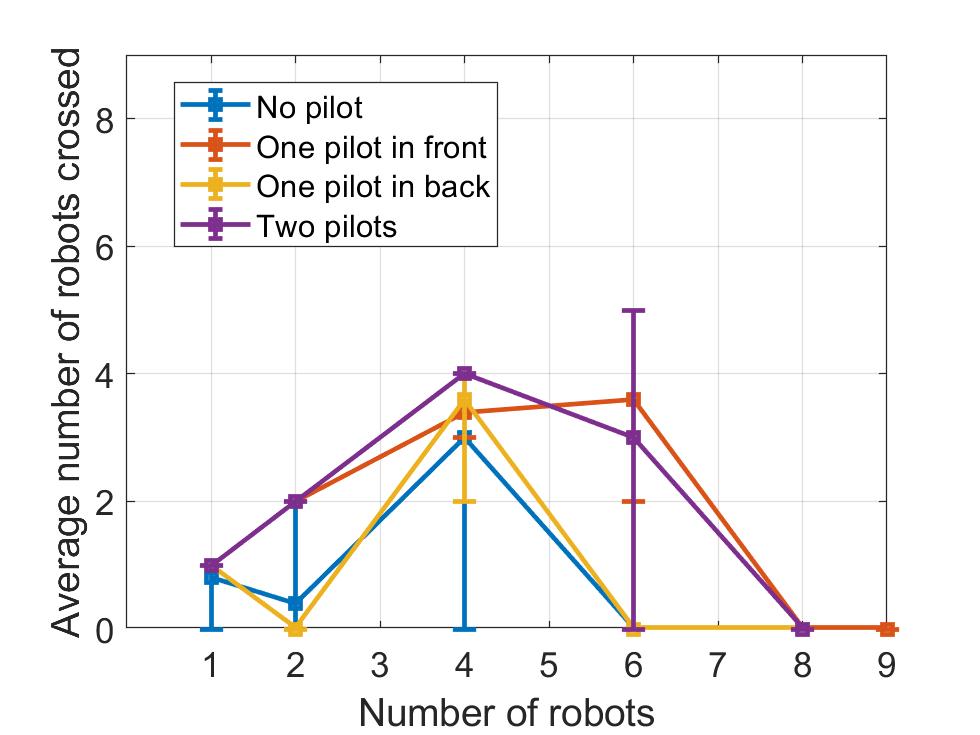}
\caption{Line configuration}\label{fig:success_line}
\end{subfigure}
\begin{subfigure}{0.235\textwidth}
\centering
\includegraphics[width=\textwidth]{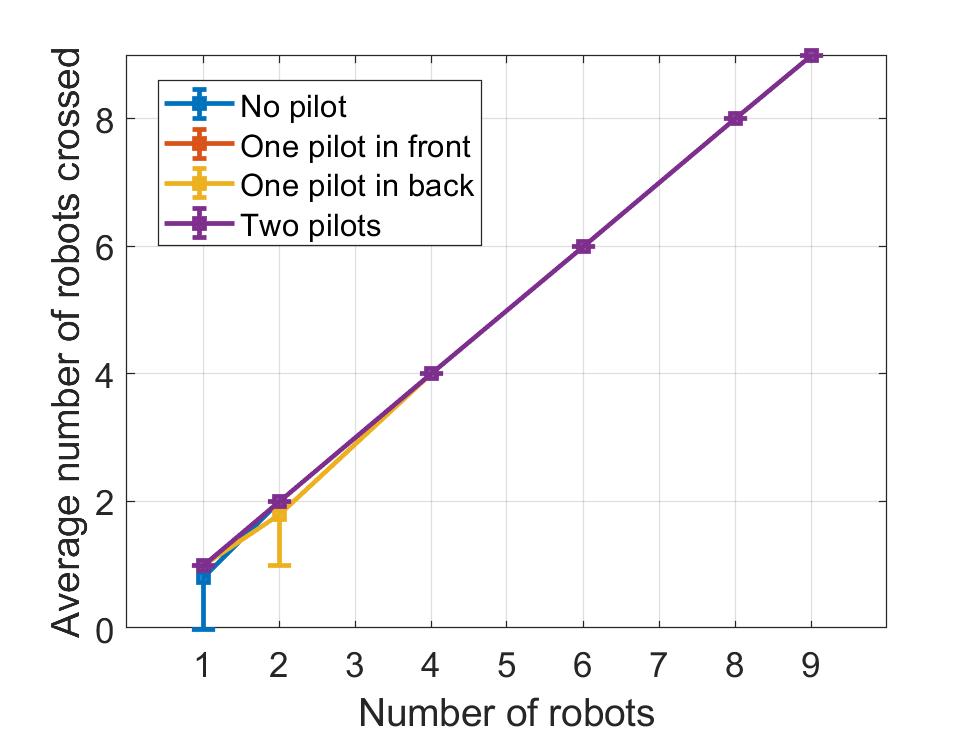}
\caption{Mesh configuration}\label{fig:success_mesh}
\end{subfigure}
\caption{Result of various number of robots crossing a gaps with 6 mm height difference, and \ang{20} heading angle. (a) (b) Maximum gap size the line/mesh configuration assembly can cross against the number of robots; (c) (d) Maximum gap size compared with the entire line/mesh assembly length, against the number of robots; (e) (f) Average number of robots that crossed a gap length = $25\%$ length of the line/mesh assembly, against the number of robots.}
\label{fig:gap_cross_experiment}
\vspace{-12pt}
\end{figure}

We performed experiments of various number of robots, with line and mesh formation, of different lengths of gaps with 6 mm height difference. The robots are coupled with a heading angle of $\ang{20}$. These two numbers are chosen based on the experiments in \cite{yi2021puzzlebots} that 6 mm is a medium height difference and $\ang{20}$ is the heading angle that gives the best performance compared with other heading angles in most of the experiments. We tested a system of 1, 2, 4, 6, 8, 9 robots with gaps lengths ranging from 10 mm to 105 mm, with 5 mm increment. Each experiment is recorded five times. We record the number of robots successfully reach the other platform. The result of the experiments is shown in Figure~\ref{fig:gap_cross_experiment}. 

Figure~\ref{fig:max_gap_line} and \ref{fig:max_gap_mesh} show that the robots in either line or mesh formation maintain the gap-crossing ability as in \cite{yi2021puzzlebots}, and can cross larger gaps with more than eight robots compared with \cite{yi2021puzzlebots}. The line formation is able to cross larger gaps with a small number of robots compared with the mesh formation. The reason is that the total length of the assembly is longer when the robots form a line compared with a mesh. As seen in Figure~\ref{fig:assembly_length_line} and \ref{fig:assembly_length_mesh}, the robots in mesh formation can cross gaps with length of larger percentage of the assembly length. For example, eight robots with two pilots are able to cross a gap $36\%$ their assembly length, while the same robots in line formation can only cross a gap $21\%$ of its assembly length. Since the coupling mechanism will introduce a small height drop of the robots, forming a mesh configuration will lock and strengthen the connection, thus giving a better performance. Robots with a pilot in the back outperform assembly without a pilot robot since the wheels of the back pilot robot provide additional pushing force for the assembly to cross a gap. A pilot robot in the front performs better compared with a back pilot since the gear wheels of the pilot robot help itself to climb onto the platform. Overall, the assembly with both front and back pilot outperforms other settings. Figure~\ref{fig:success_line} and \ref{fig:success_mesh} show the average number of robots that crossed a given gap size of $25\%$ length of the entire assembly. The error bars mark the minimum and the maximum number of robots that crossed in this setting. We observe that with the mesh configuration, all robots can cross the gap in most cases. Although in some cases, the system with two pilots line formation has less robots that crossed, the best case of the two-pilot system has more robots that crossed compared with the other settings.

\section{Conclusions and Future Works}
In this paper, we proposed a heterogeneous robot swarm system consists of pilot robots helping climb onto platforms and non-pilot robots providing flexible motions to form functional configurations. Based on our proposed system, we developed the connection-pair-oriented configuration control algorithm, enabling the robots to form various coupling configurations. Our currently approach relies heavily on the motion capture system, limiting the robots to only indoor lab environment. Further studies would include sensor integration, motions on uneven terrains, optimality and scalability analysis on the algorithm, dynamic reconfiguration, and automatic configuration generation based on different tasks and environments.

\bibliographystyle{plain}
\bibliography{ref}

\end{document}